\documentclass[10pt,twocolumn,letterpaper]{article}
\usepackage[algorithms]{wacv} 

\usepackage{graphicx}
\usepackage{amsmath}
\usepackage{amssymb}
\usepackage{booktabs}

\usepackage[utf8]{inputenc} 
\usepackage[T1]{fontenc}    
\usepackage{url}            
\usepackage{amsfonts}       
\usepackage{nicefrac}       
\usepackage{microtype}      
\usepackage[table]{xcolor}         
\usepackage{multirow}       
\usepackage{pifont}
\usepackage{float}
\usepackage{diagbox}
\usepackage[accsupp]{axessibility}  

\usepackage[pagebackref,breaklinks,colorlinks,bookmarks=false]{hyperref}

\usepackage[capitalize]{cleveref}
\crefname{section}{Sec.}{Secs.}
\Crefname{section}{Section}{Sections}
\Crefname{table}{Table}{Tables}
\crefname{table}{Tab.}{Tabs.}

\newlength\savewidth\newcommand\shline{\noalign{\global\savewidth\arrayrulewidth
\global\arrayrulewidth 1pt}\hline\noalign{\global\arrayrulewidth\savewidth}}
\newcommand{\tablestyle}[2]{\setlength{\tabcolsep}{#1}\renewcommand{\arraystretch}{#2}\centering\small}
\makeatletter\renewcommand\paragraph{\@startsection{paragraph}{4}{\z@}
{.4em \@plus1ex \@minus.2ex}{-.5em}{\normalfont\normalsize\bfseries}}\makeatother

\newcolumntype{x}[1]{>{\centering\arraybackslash}p{#1pt}}
\newcolumntype{y}[1]{>{\raggedright\arraybackslash}p{#1pt}}
\newcolumntype{z}[1]{>{\raggedleft\arraybackslash}p{#1pt}}

\definecolor{citecolor}{RGB}{34,139,34}
\definecolor{citecolor2}{HTML}{0071bc}
\definecolor{lightred}{RGB}{241,140,142}
\definecolor{defaultcolor}{gray}{0.9}
\definecolor{demphcolor}{gray}{.7}
\definecolor{linkcolor}{HTML}{ED1C24}

\newcommand{\cmark}{\ding{51}}%
\newcommand{\xmark}{\ding{55}}%

\def\wacvPaperID{*****} 
\def\confName{WACV}
\def\confYear{2024}

\title{Learning Part Segmentation from Synthetic Animals}

\author{Jiawei Peng\textsuperscript{1}\quad Ju He\textsuperscript{1}\quad Prakhar Kaushik\textsuperscript{1}\quad Zihao Xiao\textsuperscript{1}  \\  Jiteng Mu\textsuperscript{2} \quad Alan Yuille\textsuperscript{1} \vspace{.5em} \\
\textsuperscript{1}Johns Hopkins University \qquad \textsuperscript{2}UC San Diego }

\begin{document}
\maketitle

\label{sec:abstract}

\begin{abstract}
Semantic part segmentation provides an intricate and interpretable understanding of an object, thereby benefiting numerous downstream tasks. However, the need for exhaustive annotations impedes its usage across diverse object types. This paper focuses on learning part segmentation from synthetic animals, leveraging the Skinned Multi-Animal Linear (SMAL) models to scale up existing synthetic data generated by computer-aided design (CAD) animal models. Compared to CAD models, SMAL models generate data with a wider range of poses observed in real-world scenarios. As a result, our first contribution is to construct a synthetic animal dataset of tigers and horses with more pose diversity, termed Synthetic Animal Parts (SAP). We then benchmark Syn-to-Real animal part segmentation from SAP to PartImageNet, namely SynRealPart, with existing semantic segmentation domain adaptation methods and further improve them as our second contribution. Concretely, we examine three Syn-to-Real adaptation methods but observe relative performance drop due to the innate difference between the two tasks. To address this, we propose a simple yet effective method called \textbf{C}lass-\textbf{B}alanced \textbf{F}ourier \textbf{D}ata \textbf{M}ixing (CB-FDM). Fourier Data Mixing aligns the spectral amplitudes of synthetic images with real images, thereby making the mixed images have more similar frequency content to real images. We further use Class-Balanced Pseudo-Label Re-Weighting to alleviate the imbalanced class distribution. We demonstrate the efficacy of CB-FDM on SynRealPart over previous methods with significant performance improvements. Remarkably, our third contribution is to reveal that the learned parts from synthetic tiger and horse are transferable across all quadrupeds in PartImageNet, further underscoring the utility and potential applications of animal part segmentation.
\end{abstract}

\section{Introduction}
\label{sec:intro}

\begin{figure*}[h]
    \centering
    \vspace{-5mm}
    \includegraphics[width=1\textwidth]{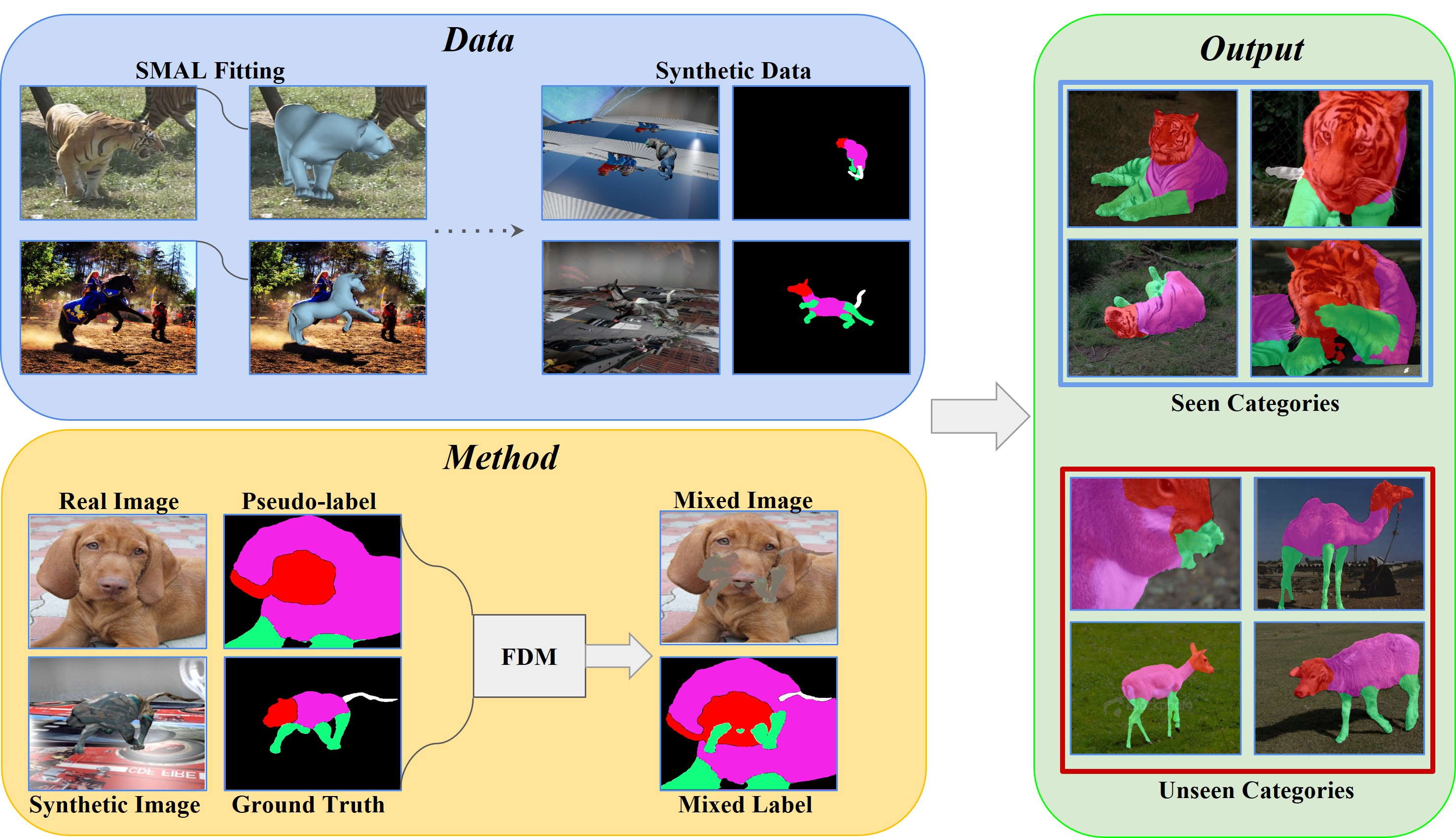}
    \vspace{-6mm}
    \caption{\textbf{Overview.} We generate synthetic animals by fitting SMAL models and rendering with random viewpoints and textures (top-left). We exploit Fourier Data Mixing (FDM) to align the spectral amplitudes of unlabeled real animals from PartImageNet and generated synthetic animals to obtain mixed images (bottom-left). Along with the Class-Balanced (CB) training strategy, our model is capable of segmenting real animals on seen categories (top-right). Moreover, the model is capable to transfer part knowledge to unseen categories (bottom-right).}
    \vspace{-3mm}
    \label{fig:teaser}
\end{figure*}

Semantic parts of an object provide a hierarchical representation which enables detailed and interpretable understanding of the object, which can facilitate various downstream tasks. For instance, humans can estimate the pose of a tiger based on the spatial configuration of its part and hence classify whether it is about to attack or lying down to rest. These hierarchical representations have also been proved to be important in many computer vision tasks, e.g., pose estimation \cite{5995741, 6909508}, detection\cite{DBLP:conf/eccv/AzizpourL12, DBLP:journals/corr/ChenMLFUY14}, segmentation \cite{NIPS2012_72b32a1f, DBLP:journals/corr/WangSLCPY15}, fine-grained recognition \cite{DBLP:journals/corr/ZhangDGD14}. However, the annotation of part segmentation on real images is very expensive, especially for general non-rigid objects, like animals. To the best of our knowledge, the only two datasets that offer animal part segmentation annotation are PASCAL-Part \cite{Pascal_Part} and PartImageNet \cite{he2022partimagenet}. While these datasets offer accurate and valuable annotations, they are limited in number of animal samples and time-consuming to scale up to more species.

By contrast, annotating parts on synthetic data is a much cheaper way to achieve the goal of scalability. Prior research \cite{mu2020learning, liu2022learning} annotated parts on 3D computer-aided design (CAD) models and rendered synthetic images based on the CAD models. With automatic-generated ground truth, this methodology offers numerous advantages, primarily in significantly reducing annotation costs. Once annotated, it can generate arbitrary number of synthetic images from arbitrary viewpoints. However, this approach comes across challenges in animal part segmentation due to the pose diversity in these CAD models is limited and does not encompass the diverse poses observed in the natural world.

As illustrated in Fig \ref{fig:teaser}, in this paper, we propose to expand the pose space for CAD data by fitting the Skinned Multi-Animal Linear (SMAL) models \cite{zuffi20173d} with more poses and utilizing them to generate supplementary synthetic data. Similar with SMPL models \cite{SMPL:2015}, SMAL models build a parametric way to represent the animal shape and pose based on strong prior and is widely used in 3D animal pose and shape estimation. This process requires additional keypoints annotation and silhouette masks to reconstruct SMAL models from images. Inspired by \cite{smal_fit1}, we replace the manual labeling process for silhouettes with the prediction of pre-trained object segmentation model \cite{li2022mask}. Combining the new SMAL data with the previous CAD data, we construct a synthetic animal dataset with diverse pose configurations of tiger and horse, termed Synthetic Animal Parts (SAP). Then we set up a new Syn-to-Real benchmark of animal part segmentation called SynRealPart from SAP to PartImageNet \cite{he2022partimagenet}, which has high-quality part segmentation annotation and provides extensive pose configurations.

To brigde the domain gap between synthetic and real, we test 3 state-of-the-art Syn-to-Real domain adaptation methods \cite{hoyer2022daformer, hoyer2022hrda, xie2023sepico} used for semantic segmentation on SynRealPart, but fail to achieve decent results. Semantic animal part segmentation is more challenging than semantic segmentation tasks because semantic parts of animals often have similar appearance and highly varying shapes. 

To address this challenge, we propose a simple yet effective method called Class-Balanced Fourier Data Mixing (CB-FDM) which consists of two parts. The first part Fourier Data Mixing (FDM) aligns the spectral amplitudes of synthetic and real images before mixing them for real domain training, thereby making the mixed images have more similar frequency content with real images. Specifically, we reconstruct the synthetic image with its original spectral phase and spectral amplitude of the real image. The reconstructed image is then mixed with the real image for training in real domain. Furthermore, we propose to use Class-Balanced Pseudo-Label Re-Weighting (CB) on certain minority class in terms of pixel frequency to alleviate the influence of the imbalanced class distribution in SAP.  

We empirically evaluate the effectiveness of our method on the SynRealPart benchmark and achieve non-trivial improvement compared to various domain adaptation methods. Specifically, we improve DAformer \cite{hoyer2022daformer} from 48.08 to 58.04 mIoU. Notably, our experiments also reveal that the learned parts from synthetic tiger and horse can be efficiently transferred (i.e. without using real labels) across all quadrupeds species in PartImageNet, even for species that have large shape variations with tiger and horse.

In summary, our main contributions are: 
\begin{enumerate}
    \item We construct a synthetic animal dataset of tigers and horses with larger diversity in pose space, named Synthetic Animal Parts (SAP), to facilitate research in animal part segmentation.
    \item We set up a new Syn-to-Real benchmark of animal part segmentation from SAP to PartImageNet and propose a simple yet effective method CB-FDM to adapt Syn-to-Real methods designed for semantic segmentation to animal part segmentation. 
    \item We reveal that the learned parts from synthetic tigers and horses are transferable across all quadrupeds in PartImageNet, which supports that core set selection for each animal category could be an effective solution to limited data, further underscoring the utility and potential applications of animal part segmentation. 
\end{enumerate}
\section{Related Work}
\label{sec:related}

\subsection{Part Segmentation}
Segmenting object parts is a long-standing problem in computer vision and there is a rich literature on the topic. The pioneering work Pictorial Structure \cite{fischler1973representation} along with following works \cite{weber2000unsupervised, felzenszwalb2005pictorial, fei2006one, zhu2007stochastic, girshick2011object} explicitly model parts and their spatial relations to the whole object. These methods share a common theme that the object-part models provide rich representations of objects and help interpretability. However, in the era of deep learning with data-driven models, research on part-based models gets hindered due to the lack of large-scale datasets.
As a result, most recent works \cite{hung2019scops,choudhury2021unsupervised,liu2021unsupervised,tritrong2021repurposing,gadre2021act,ziegler2022self,saha2022improving} mainly concentrate on unsupervised or self-supervised co-part segmentation. 
Both rigid \cite{lu2014parsing, 8237332} and non-rigid objects \cite{gong2018instancelevel, Liang_2015, wang2014semantic, 6248101} have been studied in part segmentation, but for non-rigid objects, recent works mainly focus on human part segmentation \cite{gong2018instancelevel, Liang_2015, wang2020learning, liu2018crossdomain} while there are still limited progress for animals due to the severer scarce of data.
In this work, we propose a new direction to solve animal part segmentation by utilizing synthetic data, which is much cheaper and easier to obtain compared to the expensive real data. We further explore how to transfer the models from synthetic to real in an unsupervised manner and achieve promising results.

\subsection{Synthetic Data}
Synthetic data generated by computer graphics techniques are effective for
model diagnosis \cite{DBLP:journals/corr/JohnsonHMFZG16, DBLP:journals/corr/ZhangQCHY16} and have boosted performance in many real-world application domains \cite{DBLP:journals/corr/FischerDIHHGSCB15, DBLP:journals/corr/HandaPBSC15a,9025595, mu2020learning,liu2022learning, DBLP:journals/corr/abs-1804-06516, DBLP:journals/corr/Varol0MMBLS17}. For synthetic animals, Haggag et al. \cite{animal_syn2} uses a marker-based motion-capture (MoCap) system to manually generate the animal poses which is time-consuming to generate more diverse poses. Mu et al. \cite{mu2020learning} uses 3D CAD animal models with their given animation sequences for data generation. However, the number of poses of CAD models is limited due to their animation sequences and thus is hard to scale up. We propose to use SMAL models \cite{zuffi20173d} to generate synthetic data with more diverse poses as a supplementary for the CAD synthetic data.

\subsection{Fourier Domain Bridging}
In recent years, there has been a renewed interest in using Fourier transform based methodologies in efforts to solve problems like domain adaptation \cite{yang2020fda, Yang2020PhaseCE, Huang2021RDARD}, domain generalization \cite{Xu2021AFF, kim2023domain}, domain gap reduction \cite{kumar21caft}, etc. Few works \cite{yang2020fda, kumar21caft} swap only low frequency component of the amplitude spectrum in order to learn better domain bridging features by aping target image style. Others \cite{Huang2021RDARD, kim2023domain} employ Fourier amplitude information to generate synthetic or noisy adversarial images using source domain amplitude spectrum. There also have been attempts \cite{Yang2020PhaseCE, Xu2021AFF} to preserve the phase information of an image to learn better domain bridging features - by creating images with interpolated amplitude spectrum \cite{Xu2021AFF} or mapping between the phase information of the source and target domains \cite{Yang2020PhaseCE}. Our Fourier Cross-Domain Data Mixing does not employ selective spectrum swap like \cite{yang2020fda, kumar21caft} for whole images, regularize optimization using adversarial images \cite{Huang2021RDARD, kim2023domain} or phase spectrum data \cite{Yang2020PhaseCE, Xu2021AFF} or simply interpolate between domain spectrum \cite{Xu2021AFF}. We simply utilize Fourier domain information along with spatial data mixing to help the model learn better cross-domain features. Our method does take inspiration from the simple, yet effective aforementioned works regarding utilisation of properties of Fourier transform.
\section{Synthetic Dataset}
\label{sec:data}

\begin{figure}[h]
    \vspace{-3mm}
    \centering
    \includegraphics[width=0.3\textwidth]{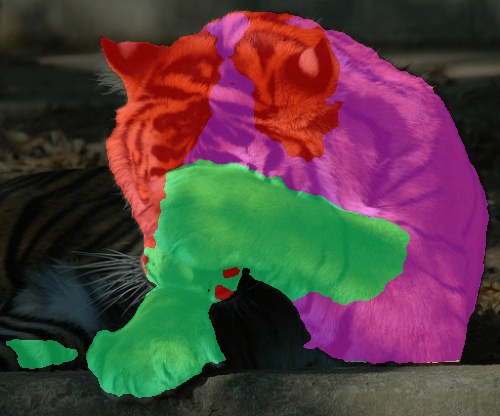}
    \vspace{-3mm}
    \caption{\textbf{Failure cases when using CAD synthetic data only.} Due to the limited poses contained in CAD models, the model fails to segment the torso out and also predicts inaccurate boundaries.}
    \label{fig:failure}
\end{figure}

By evaluating the results of utilizing CAD data for Syn-to-Real animal part segmentation on PartImageNet, we identify a few failure cases involving unusual poses (e.g. lying) and animals with self-occlusions. One typical example is shown in Fig \ref{fig:failure}.
Adding synthetic data with more unusual poses is the most intuitive solution. However, the animation sequences for each CAD animal model provide limited poses and public animal motion capture data is also scarce, which make the process difficult. 
Therefore, we opt to utilize SMAL models \cite{zuffi20173d}, which built a parametric way to represent the animal shape and pose based on strong prior. SMAL models are able to precisely reconstruct animal poses from \(2\)D images by using the animal keypoints annotation and silhouette (i.e. foreground) mask. Moreover, people only need to annotate parts for one SMAL model as these models share the same vertex IDs, which makes the data generation super efficient. We present the details of our data generation process and statistics below.

\subsection{Data Generation}
\noindent \textbf{(\(1\)) 2D Annotation.} Firstly, we select several natural animal images depicting poses not including in animation sequences of CAD models from online resources. Secondly, we carefully annotate 26 keypoints per image based on the original keypoints definition of SMAL model. Inspired by \cite{smal_fit1}, we replace the manual labeling process for silhouettes with the prediction of pre-trained object segmentation model\cite{li2022mask}.

\noindent \textbf{(\(2\)) SMAL Fitting.} With the keypoints and silhouette masks, we utilize SMALR \cite{Zuffi:CVPR:2018} which recovers refined 3D models from 2D images. For detailed fitting process, we refer to \cite{Zuffi:CVPR:2018} for more information.

\noindent \textbf{(\(3\)) 3D Part Annotation.} 
The annotation of 3D parts is done by grouping the vertex IDs of each semantic part. Firstly, we select an SMAL model of arbitrary animal in a typical pose. Secondly, we use Blender \cite{blender} to group the vertex IDs of each part and save them. Since all SMAL models share the same vertex IDs, this part annotation can be directly applied to other SMAL models. 

\noindent \textbf{(\(4\)) Rendering Images and Part Segmentation Masks.} 
Following previous work \cite{mu2020learning, liu2022learning}, we use Blender as our render and randomize render parameters (e.g., viewpoint, lighting, and object texture) to promote domain generalization. The background images are randomly sampled from COCO \cite{DBLP:journals/corr/LinMBHPRDZ14}. 
The 2D part segmentation mask is obtained through directly projecting the annotated parts in the third step.

\subsection{Dataset Statistics}
We create our SMAL data utilizing the aforementioned pipeline. Specifically, we generate a total of 4,400 synthetic tiger images from 11 distinct poses. Each pose encompasses 100 viewpoints and 4 transformations, notably rotations. Similarly, we produce 2,000 synthetic horse images derived from 10 poses, with 100 viewpoints and 2 transformations. All animals are rendered with randomly selected textures from real images.
It is important to highlight that our SMAL synthetic data offers a broad range of unusual poses, including lying, climbing, and other movements beyond walking and running, which are not included in the CAD synthetic data introduced in \cite{mu2020learning}.

We further integrate our SMAL synthetic data with the existing CAD synthetic data \cite{mu2020learning}, creating a comprehensive dataset specifically designed for animal part segmentation, named Synthetic Animal Parts (SAP). This combined dataset encompasses a total of 14,400 images for tigers and 12,000 images for horses. SAP offers diverse pose configurations for both tigers and horses, accompanied by accurate part masks. We believe that SAP will serve as valuable resources for advancing research in animal part segmentation.
\section{Methods}
\label{sec:method}
In this section, we begin by presenting the formulation for syn-to-real part segmentation. Subsequently, we introduce the intuition and details of Fourier Data Mixing (FDM). Finally, we illustrate the motivation behind Class-Balanced Pseudo-Label Re-Weighting (CB).  

\subsection{Preliminaries}
\paragraph{Syn-to-Real Part Segmentation} Similar to semantic segmentation, a part segmentation model predicts the pixel-wise label for parts of a object where each part is a category. For example, in our quadruped animals setup, the part classes are head, torso, leg, tail of a quadruped animal. We denote the source domain as \(\mathcal{D}_s = \{(x_s^{(i)}, y_s^{(i)}) \}_{i=1}^{N_s}\) with \(N_s\) samples drawn from the synthetic domain, where \(x_s^{(i)} \in X_s\) is an image, \(y_s^{(i)} \in Y_s\) is the corresponding pixel-wise one-hot label over \(K+1\) classes (including background). Note that \(K\) is the number of part classes. Similarly, the unlabeled target domain is denoted as \(\mathcal{D}_t = \{x_t^{(i)} \}_{i=1}^{N_t}\) with \(N_t\) samples drawn from the real domain. This work aims to learn a part segmentation model that can effectively transfer part knowledge from the synthetic domain to the real domain. In addition, the part segmentation model is also assumed to have ability to transfer parts from one object class to a similar object class.

\subsection{Fourier Data Mixing}
Recent unsupervised syn-to-real translation methods \cite{hoyer2022daformer, hoyer2022hrda, hoyer2023mic, xie2023sepico, chen2022deliberated} use self-training (i.e. using pseudo labelled real images for training on target domain) framework. The quality of pseudo-labels for the target images is crucial for achieving satisfactory convergence. As a solution, \cite{dacs} proposed mixing the source and target domain image patches with binary masks obtained using various mixing algorithms \cite{olsson2020classmix, cutmix}. 
On the other hand, spectral information obtained through Fourier transform can be utilized to provide a global description of the image\cite{amplitude} as well as help in learning domain bridging features \cite{Yang2020PhaseCE, Xu2021AFF}.
Inspired by both of these concepts, we align the global information of the mixed regions to help the model learn better cross-domain features by aligning the spectral amplitudes of the mixed images. 
Let \(\mathcal{F}^{A} \), \( \mathcal{F}^{P}\) \(: \mathbb{R}^{H \times W \times 3} \rightarrow \mathbb{R}^{H \times W \times 3}\) be the amplitude and phase components of the Fourier transform \(\mathbf{F}\) of an RGB image, i.e., for a single channel image \(x\) we have: 
\begin{equation} \label{eq5}
\mathcal{F}(x)(m,n) = \sum_{h,w} x(h, w) e^{-j2\pi (\frac{h}{H}m + \frac{w}{W}n)}, j^2 = -1
\end{equation}

, which can be implemented efficiently use the Fast Fourier Transform(FFT) algorithm\cite{fft}. \(\mathcal{F}^{-1}\) is the inverse Fourier transform that maps the spectral signals back to the image space. Then in the mixed sampling stage, given two random samples \( (x_s, y_s) \sim \mathcal{D}_s, x_t \sim \mathcal{D}_t\), we use FFT to get the spectral signals(amplitude and phase) from both, then the Fourier alignment can be formulated as:
\begin{equation} \label{eq6}
x_{mixed} = \mathbf{M} \odot \mathcal{F}^{-1}(\mathcal{F}^A(x_t), \mathcal{F}^P(x_s)) + (\mathbf{1} - \mathbf{M}) \odot x_t
\end{equation}
\begin{equation} \label{eq3}
y_{mixed} = \mathbf{M} \odot y_s + (\mathbf{1} - \mathbf{M}) \odot \hat{y}_t
\end{equation}
 , where \( \mathbf{M}\) denotes a binary mask generated by ClassMix \cite{olsson2020classmix}, indicating which pixel needs to be copied from the augmented source domain
and pasted to the target domain, \(\mathbf{1}\) is a mask filled with ones, and \(\odot\) represents the element-wise multiplication operation, and \(\hat{y}_t\) is the pseudo-label of \(x_t\). In the formulation, the amplitude of the source image \(\mathcal{F}^A(x_s)\) is replaced by that of the target image \(x_t\). Then the modified spectral representation of \(x_s\), with its phase component unchanged, is mapped back the image domain to get the augmented image. We hypothesize that by making the source region of the mixed images have more similar frequency content with the target region, it will be harder for the model the learn the difference between the source and target regions and force the model to learn more domain-invariant features, thereby achieving better performance. 

\subsection{Class-Balanced Pseudo-Label Re-Weighting}

Our synthetic dataset exhibiting a class imbalanced distribution in terms of pixel frequency \ref{ref:tab_method_pixel}. Although existing methods \cite{hoyer2022daformer, hoyer2022hrda, hoyer2023mic, xie2023sepico} employ the Rare Class Sampling (RCS) training strategy to sample the source images which contain minority classes in terms of pixel frequency more often, the supporting gradients \cite{zhang2023deep} for some minority classes may still be very limited at the early training stage. It arises due to the fact that many images containing minority classes in SAP marginally surpass the pre-defined threshold of pixel number in RCS. To prevent RCS from sampling in a limited range of images, we are not able to set a high threshold which help the sampled images to have relatively higher pixel frequency for the minority class. Therefore, even one image containing certain minority class is sampled by RCS, the supporting gradients may still be limited.
In the research of semi-supervised learning (SSL) in the context of class-imbalanced data for the classification task, there is an observation that the undesired performance of existing SSL algorithms on imbalanced data is mainly due to low recall on minority classes in terms of number of samples, but the precision on minority classes is surprisingly high \cite{wei2021crest}.
We observe a similar phenomenon w.r.t animal head part (shown in supplementary), which is one of the minority classes in terms of pixel frequency in our synthetic data. The predictions of the animal head parts for real images hardly give false positive results but mainly the false negative. Therefore, in order to boost pseudo-label confidence for the head part, we give it more weight.
We name the multiplicative factor of the pseudo-label weights of animal head as \(\beta \) for convenience. This technique mitigates the issue that the model may receive very limited supporting gradients for animal head class at the early stage of training, leading to unsatisfactory performance.

\begin{table}[]
    \caption{\textbf{Pixel frequency of 4 classes in the Synthetic Animal Parts (SAP) dataset.} We compute the statistics of pixel frequency in SAP which exhibits the class-imbalance distribution.}
    \begin{tabular}{c|cccc}
    class           & head   & torso  & leg    & tail  \\ \shline
    pixel frequency & 12.7\% & 55.6\% & 25.9\% & 5.8\%
    \end{tabular}
    \label{ref:tab_method_pixel}
\end{table}

\section{Experiments}
\label{sec:exp}
In this section, we first provide our implementation details including how we construct train and test set on SynRealPart and training settings in Sec. \ref{ref:sec_exp_data}. After setting the stage, we introduce our main results, compared with state-of-the-art methods in Sec. \ref{ref:sec_exp_main}, followed by ablation studies in Sec. \ref{ref:sec_exp_ablation} to validate the key designs in our model. In the end, we further explore the part knowledge transfer in Sec. \ref{ref:sec_exp_transfer} that part segmentation results are transferable among species of similar structures regardless of shape and texture difference which points to a promising future direction. Qualitative visualization results are presented as well. 

\subsection{Implementation Details}
\label{ref:sec_exp_data}
\noindent \textbf{Data} For synthetic data, we utilize 23520 images (11520 synthetic tiger images + 12000 synthetic horse images) in SAP for training. For real training data, we select 5942 quadrupeds images from PartImageNet \cite{he2022partimagenet} which excludes tiger images. Note that PartImageNet doesn't have horse images. Then we can get a UDA setting that all tiger and horse information are learned from our synthetic data. We select another 1213 quadrupeds images as our main test set which includes tiger images.

\noindent \textbf{Training settings.} We conduct our experiments using SegFormer \cite{segformer}, DAFormer \cite{hoyer2022daformer}, HRDA \cite{hoyer2022hrda} and SePiCo \cite{xie2023sepico}. MiT-b5 (Mix Transformer encoders) pre-trained on ImageNet-1k is adopted as the backbone for above methods. If not specified, we train all our models with batch size 2 on a single GPU for 30k iterations. We set the learning rate of the decoder head to be 6e-4 and the backbone has a learning rate multiplier 0.1. We use AdamW \cite{adamw} optimizer with weight decay 0.01. For data augmentation, we adopt random color jittering. The input image is cropped into $512 \times 512$ ($1024 \times 1024$ for HRDA).

\subsection{Main Results}
\label{ref:sec_exp_main}

\begin{table*}[]
\centering
\tablestyle{9pt}{1.05}
\caption{\textbf{Comparison of part segmentation (mIoU) on our test set.} CB-FDM brings non-traivial improvements to all the syn-to-real methods, especially for DAFormer \cite{hoyer2022daformer}, which makes DAFormer + CB-FDM the optimal syn-to-real solution in our current benchmark in terms of mIoU. Numbers are averaged over 3 random seeds. *: train SegFormer with synthetic data and real data successively for each iteration (equivalent to replacing pseudo-labels with real labels in DAFormer). ``bg" stands for background.}
\label{tab_exp_tiger+horse}
\begin{tabular}{c|c|ccccc|c}
data & method & head & torso & leg & tail & bg & mIoU \\ \shline
SAP &  & 49.71 & 38.84 & 31.86 & 7.53 & 83.11 & 42.21 \\ \cline{1-1}
PartImageNet\cite{he2022partimagenet} & \multirow{-2}{*}{SegFormer\cite{segformer}} & 85.79 & 72.69 & 60.00 & 58.27 & 96.81 & 74.71 \\ \hline
 & DAFormer\cite{hoyer2022daformer} & 44.94 & 48.66 & 44.11 & 12.21 & 90.45 & 48.08 \\
 & \cellcolor{defaultcolor}DAFormer + CB-FDM & \cellcolor{defaultcolor}69.40 & \cellcolor{defaultcolor}56.81 & \cellcolor{defaultcolor}49.17 & \cellcolor{defaultcolor}21.42 & \cellcolor{defaultcolor}93.39 & \cellcolor{defaultcolor}58.04 \\
 & HRDA\cite{hoyer2022hrda} & 44.91 & 37.16 & 36.51 & 12.75 & 88.00 & 43.87 \\
 & \cellcolor{defaultcolor}HRDA + CB-FDM & \cellcolor{defaultcolor}64.21 & \cellcolor{defaultcolor}40.69 & \cellcolor{defaultcolor}42.03 & \cellcolor{defaultcolor}6.54 & \cellcolor{defaultcolor}88.21 & \cellcolor{defaultcolor}48.34 \\
 & SePiCo\cite{xie2023sepico} & 55.67 & 57.83 & 42.53 & 15.6 & 91.29 & 52.58 \\
\multirow{-6}{*}{\begin{tabular}[c]{@{}c@{}}SAP + Unlabeled\\ PartImageNet\end{tabular}} & \cellcolor{defaultcolor}SePiCo + CB-FDM & \cellcolor{defaultcolor}58.72 & \cellcolor{defaultcolor}52.71 & \cellcolor{defaultcolor}49.37 & \cellcolor{defaultcolor}21.17 & \cellcolor{defaultcolor}92.27 & \cellcolor{defaultcolor}54.85 \\ \hline
SAP + PartImageNet & SegFormer* & 86.33 & 73.37 & 62.30 & 56.65 & 96.72 & 75.08
\end{tabular}
\end{table*}

\begin{figure*}[h]
\centering
\includegraphics[width=1\textwidth]{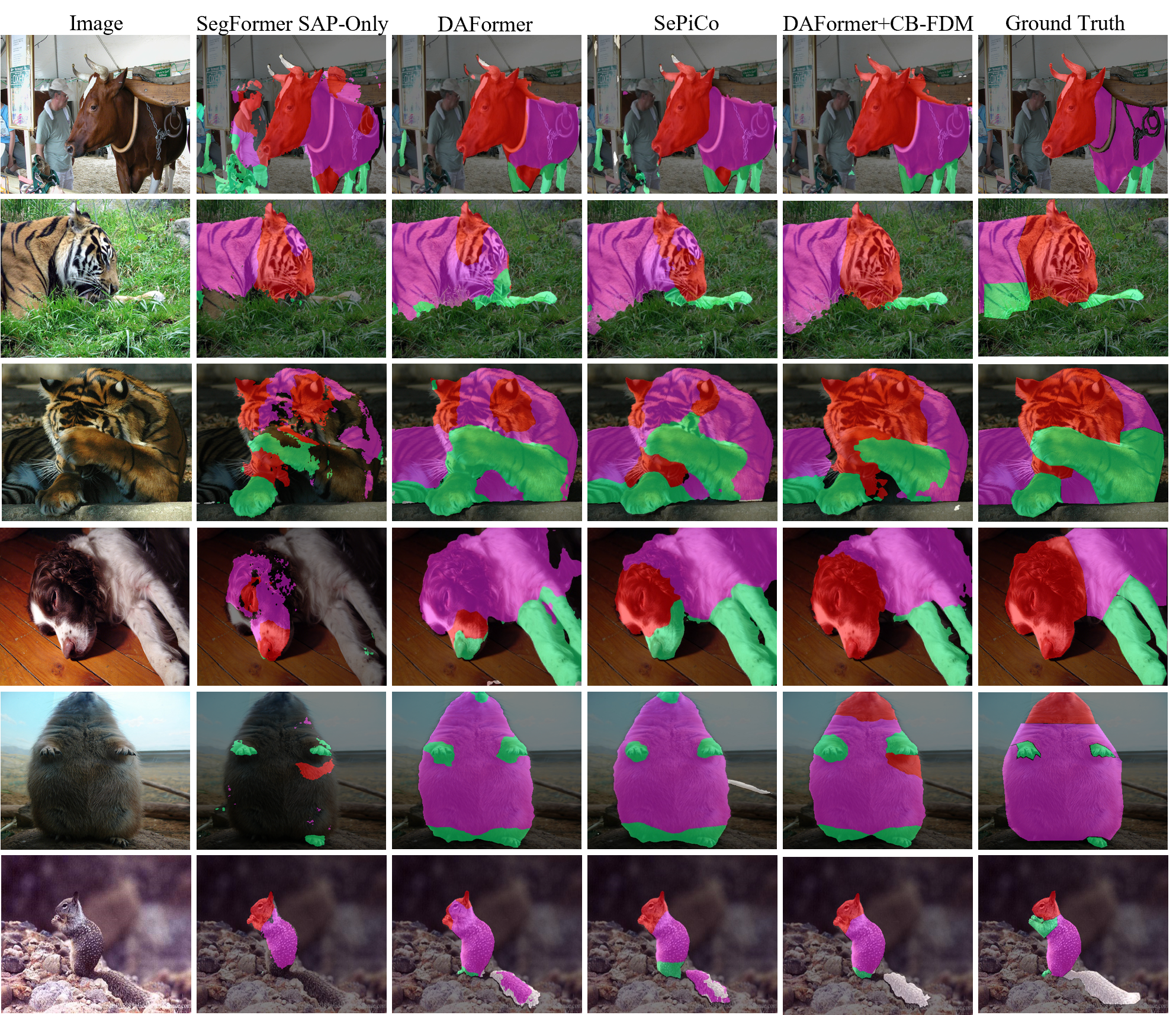}
\caption{\textbf{Qualitative comparisons for different methods on our test set.} Note that our CB-FDM with DAFormer \cite{hoyer2022daformer} produces more accurate part segmentation results in challenging poses (e.g., row 2\&3\&4) and unseen species that have large shape difference with tiger and horse (e.g., row 5\&6).}
\label{ref:vis_exp_main}
\end{figure*}

\begin{table*}[ht]
    \centering
    \tablestyle{12pt}{1.05}
    \caption{\textbf{CB-FDM ablation studies on our test set in terms of mIoU.} FDM \& CB both provide non-trivial improvement based on DAFormer. Numbers are averaged over 3 random seeds.}
    \begin{tabular}{c|cc|ccccc|c}
        method & FDM & CB & head & torso & leg & tail & bg & mIoU \\ \shline
        \multirow{4}{*}{DAFormer \cite{hoyer2022daformer}} & \xmark & \xmark & 44.94 & 48.66 & 44.11 & 12.21 & 90.45 & 48.08  \\
         & \cmark & \xmark & 50.14 & 50.72 & 49.17 & 20.74 & 93.11 & 52.77 \\
         & \xmark & \cmark & 71.69 & 56.21 & 46.37 & 16.15 & 92.19 & 56.52 \\
         & \cmark & \cmark & \cellcolor{defaultcolor}69.40 & \cellcolor{defaultcolor}56.81 & \cellcolor{defaultcolor}49.17 & \cellcolor{defaultcolor}21.42 &
         \cellcolor{defaultcolor} 93.39 &
         \cellcolor{defaultcolor}\textbf{58.04}
    \end{tabular}
    \label{ref:tab_exp_module}
\end{table*}

\begin{table*}[]
    \centering
    \tablestyle{14pt}{1.05}
    \caption{\textbf{Parameter study of the class balance weight $\beta$.} We achieve the optimal performance when $\beta = 2$. Numbers are averaged over 3 random seeds.}
    \begin{tabular}{c|c|ccccc|c}
    method & $\beta$ & head & torso & leg & tail & bg & mIoU \\ \shline
    \multirow{4}{*}{DAFormer \cite{hoyer2022daformer} + CB} & 1.5 & 58.97 & 52.89 & 41.87 & 19.54 & 92.17 & 53.09 \\
     & \cellcolor{defaultcolor}2 & \cellcolor{defaultcolor}71.69 & \cellcolor{defaultcolor}56.21 & \cellcolor{defaultcolor}46.37 & \cellcolor{defaultcolor}16.15 &  \cellcolor{defaultcolor}92.19 &\cellcolor{defaultcolor}\textbf{56.52} \\
     & 2.5 & 71.72 & 57.74 & 45.86 & 13.67 & 92.46 & 56.29 \\
     & 3 & 72.68 & 59.86 & 45.79 & 8.32 & 92.04 & 55.74
    \end{tabular}
    \label{ref:tab_exp_beta}
\end{table*}

\begin{table*}[]
\centering
\tablestyle{6pt}{1.05}
\caption{\textbf{Ablation study for adding SMAL synthetic data.} Introducing SMAL data lead to a non-trivial improvement. Numbers are averaged over 3 random seeds.}
\vspace{-2mm}
\begin{tabular}{c|cc|ccccc|c}
\multirow{2}{*}{Method} & \multicolumn{2}{c|}{synthetic data source} & \multirow{2}{*}{head} & \multirow{2}{*}{torso} & \multirow{2}{*}{leg} & \multirow{2}{*}{tail} & \multirow{2}{*}{bg} & \multirow{2}{*}{mIoU} \\ \cline{2-3}
 & \textit{CAD} & \textit{SMAL} &  &  &  &  &  &  \\ \shline
\multirow{2}{*}{SegFormer\cite{segformer}} & \cmark  & \xmark  & 44.51 & 35.72 & 31.45 & 8.20 & 83.19 & 40.61 \\
 & \cmark  & \cmark & \cellcolor{defaultcolor}49.71 & \cellcolor{defaultcolor}38.84 & \cellcolor{defaultcolor}31.86 & \cellcolor{defaultcolor}7.53 & \cellcolor{defaultcolor}83.11 & \cellcolor{defaultcolor}\textbf{42.21} \\ \hline
\multirow{2}{*}{\begin{tabular}[c]{@{}c@{}}DAFormer\cite{hoyer2022daformer} \\+ CB-FDM\end{tabular}} & \cmark  & \xmark  & 59.09 & 45.40 & 48.88 & 31.98 & 91.90 & 55.45 \\
 & \cmark  & \cmark & \cellcolor{defaultcolor}69.40 & \cellcolor{defaultcolor}56.81 & \cellcolor{defaultcolor}49.17 & \cellcolor{defaultcolor}21.42 & \cellcolor{defaultcolor}93.39 & \cellcolor{defaultcolor}\textbf{58.04}
\end{tabular}
\label{ref:tab_exp_data}
\end{table*}

\begin{table*}[]
\centering
\tablestyle{11pt}{1.05}
\caption{\textbf{Part segmentation results (mIoU) of \textit{Horse} and \textit{Tiger} settings.} \textit{Horse} setting is trained with synthetic horse and horse-like real images, and then test on tiger-like real images. Similarly, \textit{Tiger} setting is trained with synthetic tiger and tiger-like real images, and then test on horse-like real images. Numbers are averaged over 3 random seeds.}
\vspace{-3mm}
\begin{tabular}{c|c|ccccc|c}
settings & method & head & torso & leg & tail & bg & mIoU \\ \shline
\textit{Tiger} & \multirow{2}{*}{\begin{tabular}[c]{@{}c@{}}DAFormer  + CB-FDM\end{tabular}} & 32.97 & 57.52 & 46.46 & 2.14 & 94.10 & 46.64 \\ \cline{1-1} \cline{3-8} 
\textit{Horse} &  & 71.22 & 54.63 & 38.26 & 12.07 & 90.11 & 53.26
\end{tabular}
\label{ref:tab_exp_transfer}
\end{table*}

Table \ref{tab_exp_tiger+horse} summaries our results on our main test set. SegFormer \cite{segformer} supervisedly trained on SAP achieves 42.21 mIoU on the test set. We observe that naively applying state-of-the-art semantic segmentation syn-to-real methods sometimes can not bring significant improvement (i.e. from 42.21 to 43.87 in terms of mIoU for HRDA \cite{hoyer2022hrda}). Notably, after applying our proposed Class-Balanced Fourier Data Mixing (CB-FDM), all syn-to-real methods get non-trivial improvement. For DAFormer, the improvement gain is up to 9.96 mIoU while we also improve HRDA by 4.47 and SePiCo \cite{xie2023sepico} by 2.27 in terms of mIoU. SePiCo\cite{xie2023sepico} which combines DAFormer \cite{hoyer2022daformer} with contrastive learning shows the optimal performance in direct application but not very sensitive to CB-FDM. We hypothesis that it is because CB-FDM is designed for the cross-entropy losses of the mixed images while SePiCo additionally has one contrastive loss for source images with their groundtruth and one contrastive loss for real images with their pesudo-labels, which mitigate the influence of CB-FDM. 
The syn-to-real results also reveal that part knowledge can be efficiently transferred among objects with similar structures regardless of shape and texture difference. With unlabeled real data and proper algorithms, the part knowledge of only synthetic tiger and horse can be much better adapted to all 46 quadrupeds in PartImageNet (i.e. DAFormer + CB-FDM improves the performance by 15.83 mIoU compared to training on SAP). We believe this finding can motivate the exploration in animal part segmentation since people will only need to select a core set of animals species in each animal category for training. In addtion, we supervisedly train a SegFormer on real training data from PartImageNet\cite{he2022partimagenet} and achieve 74.71 mIoU. When we supervisedly train on both synthetic and real data, there is another 0.37 improvement in terms of mIoU. 

\noindent \textbf{Visualizations.} We also conduct qualitative comparison as illustrated in Fig. \ref{ref:vis_exp_main}. We show that models trained only on SAP usually fail to generalize to real images while naively adapting semantic segmentation syn-to-real methods yields many incorrect part predictions. On the contrary, our CB-FDM with DAFormer is able to predict more accurate boundary for different parts even in the scenarios of challenging poses (e.g. row 2\&3\&4) and unseen species that have large shape difference with tiger and horse (e.g. row 5\&6).

\subsection{Ablation Study}
\label{ref:sec_exp_ablation}
\noindent \textbf{Effectiveness of proposed modules.}
Table \ref{ref:tab_exp_module} summarizes the effects of the key designs in our method. We note that after applying Fourier Data Mixing (FDM), we can obtain a general improvement on all classes which lead to an overall improvement of 4.69 on mIoU. Class-Balanced (CB) sampling brings a significant improvement on head class (i.e. 26.75 mIoU) as it pays more attention to it while also improves the performance on all the other classes. When combining these two parts together, we obtain our final method Class-Balanced Fourier Data Mixing (CB-FDM) and achieves 58.04 mIoU. We can notice slight performance drop on head but improvements on all the other classes compared to using CB only. We assume that FDM prevents the over-fitting on head when using CB and thus is a good combination with it.

\noindent \textbf{Influence of balance strength $\beta$.} Tab. \ref{ref:tab_exp_beta} presents our results on controlling the balance strength through $\beta$. As can be observed, a reasonable strong balance weight (i.e. $\beta$ > 1.5) is required to achieve good results, while setting it too large will also harm the performance as well. We set $\beta$ to be 2 as our default setting according to this experimental results.

\noindent \textbf{Synthetic Data Source.} Tab. \ref{ref:tab_exp_data} shows ablation studies on using our SMAL synthetic data. As we can observe from the comparisons, after introducing our SMAL synthetic data, SegFormer(i.e. synthetic only) achieves 1.6 improvement while DAFormer + CB-FDM achieves 2.59 improvement in terms of mIoU. However, we notice a performance drop on tail after introducing SMAL data. We hypothesis that one potential reason is the inaccurate tail shape for SMAL fitting algorithms when tails are self-occluded for unusual poses in real images.

\subsection{Zero-Shot Part Knowledge Transfer}
\label{ref:sec_exp_transfer}
In Sec. \ref{ref:sec_exp_main}, we already discussed that Table \ref{tab_exp_tiger+horse} implies the part knowledge from synthetic tiger and horse can be efficiently transferred to all quadrupeds in PartImageNet. To further explore the power of this transfer ability, we design 2 more zero-shot settings for only tiger and horse respectively because they have relatively large shape difference. Since PartImageNet doesn't have horse classes, we extend the standard to tiger-like and horse-like animals. We select a tiger-like (i.e. tiger, cheetah, lion) animal set which includes 630 images and a horse-like (i.e. goat, deer, buffalo, etc) animal set which includes 1016 images from PartImageNet. Then we have the following 2 unsupervised syn-to-real settings: 1) \textit{Tiger}: Train on synthetic tiger and tiger-like set, and test on horse-like set; 2) \textit{Horse}: Train on synthetic horse and horse-like set, and test on tiger-like set. From table \ref{ref:sec_exp_transfer}, we can see both settings can transfer the torso knowledge pretty good as that is the part which have the most similar shape. We assume the huge difference in head and tail performance between these 2 settings is mainly caused by the ambiguity problem of tail and horn (refer to supplementary for its visualization). Horn is an unseen parts for our synthetic data. Our real groundtruth regard it as a part of head while the model trained on synthetic often predicts it as tail. Horse-like set mainly consists of animals with horns,
which leads to a huge performance drop on head and tail for \textit{Tiger} setting. Without the ambiguity issue, we can see the \textit{Horse} setting has achieved 53.26 mIoU even with a tiny amount of unlabeled real data and this zero-shot setup. We believe these findings point to a promising future direction in solving limited data problems in animal part segmentation and constructing training data more efficiently (i.e. core set selection for each animal category).

\section{Conclusion}
\label{sec:conclusion}
In this paper, we propose to use SMAL models for efficiently generating synthetic data with diverse pose configurations and further construct a synthetic animal dataset of tigers and horses with part segmentation groundtruth termed as Synthetic Animal Parts (SAP). Then we set up a new Syn-to-Real benchmark of animal part segmentation from SAP to PartImageNet called SynRealPart, and we propose a simple yet effective method called Class-Balanced Fourier Data Mixing (CB-FDM) consisting of Fourier Data Mixing (FDM) and Class-Balanced Pseudo-label Re-weighting (CB) to improve the performance of existing unsupervised syn-to-real adaptation methods designed for semantic segmentation on it. Our experiments also reveal that the learned parts from synthetic tiger and horse are transferable across all quadrupeds in PartImageNet, which supports that core set selection for each animal category could be an effective solution to limited data, further underscoring the utility and potential applications of animal part segmentation.

\noindent \textbf{Limitations.} At present, our SAP dataset focuses exclusively on two animal species: tigers and horses, which are two representative animals for quadrupeds. For quadrupeds, we plan to add one or two quadrupeds which have horns to solve the ambiguity problems between horn and tail. We also plan to expand our synthetic animal data to contain more animal categories like bird and reptile. While SMAL models are for quadrupeds only, we need to explore other efficient solutions to creating diverse poses, but using animation sequences of CAD models for normal poses and using 3D models reconstructed from real images for unusual poses may still be our core strategy. Furthermore, note that we cannot apply CB to the "tail" class since its prediction does not have a high precision like the "head" class (shown in supplementary). One reason for it is the ambiguity problem between tail and horn which may be solved by adding synthetic animals with horns. However, tail is still the hardest part to segment since it has the biggest shape and deformation variance among different animals. Improving segmentation accuracy on tail is quite challenging and remains unsolved. Lastly, while our model demonstrates the ability to transfer part knowledge across different animal species, it still lacks the capability to handle unseen parts (i.e. horns).

{\small
\bibliographystyle{unsrt}
\bibliography{egbib}
}

\clearpage

\noindent\textbf{\Large Supplementary Material}

\makeatletter
\renewcommand{\theHsection}{papersection.\number\value{section}} 
\renewcommand{\thesection}{\Alph{section}}
\renewcommand{\thefigure}{S\arabic{figure}}
\renewcommand{\thetable}{S\arabic{table}}
\setcounter{section}{0}

\setcounter{figure}{0}
\setcounter{table}{0}
\makeatother

\section*{Supplementary Material}
\section{More Implementation Details}
We disable the Thing-Class ImageNet Feature Distance(FD) \cite{hoyer2022daformer} for all methods \cite{hoyer2022daformer, hoyer2022hrda} containing it. It is a regularization technique that uses ImageNet features which are trained from objects to provide guidance to segment object classes, which is inappropriate for segmenting semantics parts of object. In addition, we use the distribution-aware pixel contrast for SePiCo \cite{xie2023sepico}.

The tiger-like animals set contains the following classes from PartImageNet \cite{he2022partimagenet}: n02129604, n02125311, n02128385, n02130308. The horse-like animals set contains the following classes: n02403003, n02415577, n02423022, n02408429, n02412080, n02422699, n02437312, n02422106, n02417914.

The implementation code and synthetic dataset will be available at {\footnotesize\url{https://github.com/RyougiJarvis/SynAnimals}}.

\section{Observations for Pseudo Labels on Head and Tail}
Table \ref{ref:tab_exp_class_stats} shows the recall of head is low while the precision is surprisingly high, which supports our claims that predictions on head tend to give false negative results and hardly give false positive. Although tail has a low recall, its precision is also very low and thus we cannot give more confidence on its prediction. This is why we can only apply our pseudo label re-weighting strategy to head class only.

\begin{table}[h]
\centering
\resizebox{\linewidth}{!}{
\begin{tabular}{c|c|cc|cc}
\multirow{2}{*}{Method} & \multirow{2}{*}{Iterations} & \multicolumn{2}{c|}{head} & \multicolumn{2}{c}{tail} \\ \cline{3-6} 
 &  & recall & precision & recall & precision \\ \shline
\multirow{3}{*}{DAFormer} & 1000 & 13.03 & 84.86 & 17.20 & 10.48 \\
 & 2000 & 16.96 & 90.34 & 26.60 & 20.43 \\
 & 3000 & 20.51 & 91.14 & 21.67 & 25.65
\end{tabular}}
\caption{The precision and recall of the predictions on target domain training data (i.e pseudo labels) at the early stage of training. The setting is the same as the our DAFormer baseline in Table \ref{ref:sec_exp_main}. }
\label{ref:tab_exp_class_stats}
\end{table}

\section{Random Texture for SMAL Synthetic Data}
\begin{figure*}[h]
\centering
\includegraphics[width=1\textwidth]{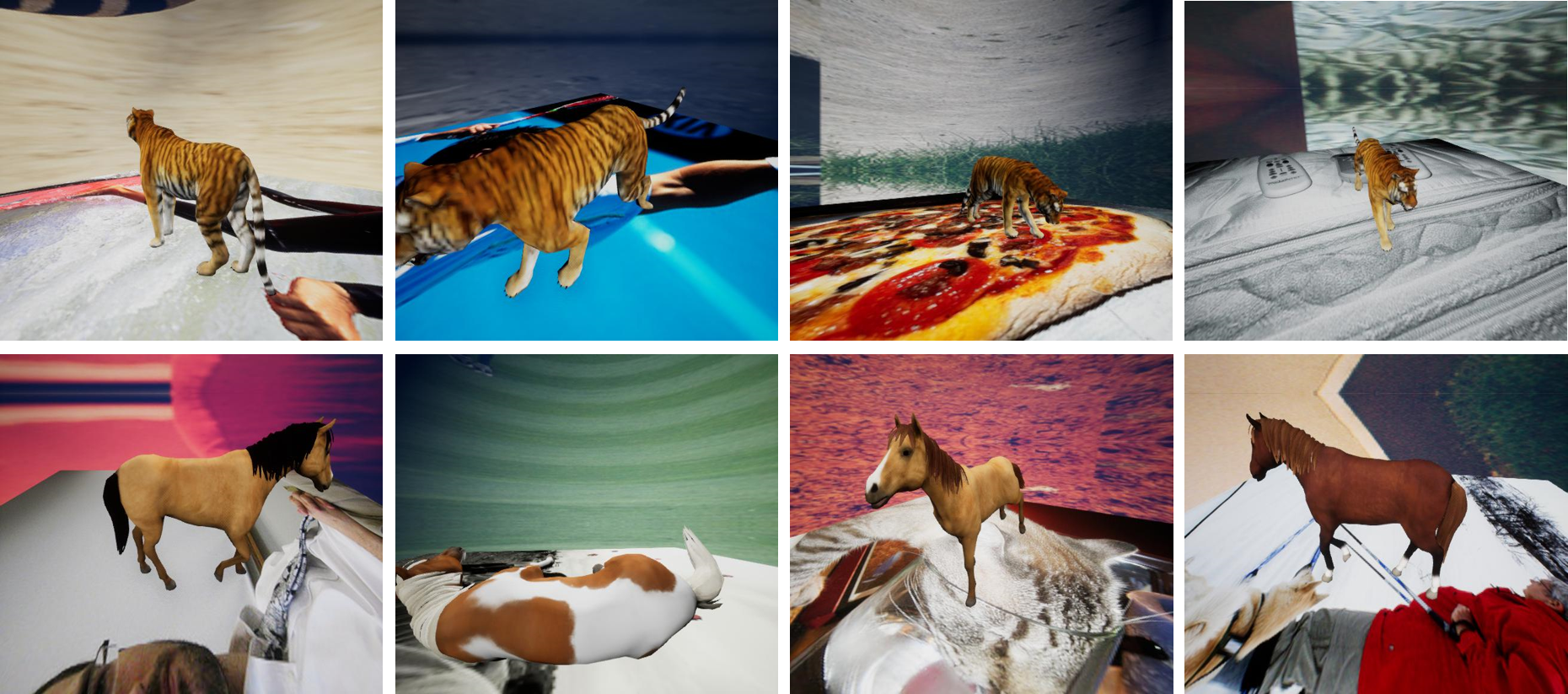}
\caption{\textbf{Real Texture Provided in CAD Synthetic Data.}}
\label{ref:vis_texture}
\end{figure*}

In the CAD synthetic data, half of the synthetic images have real textures provided by the CAD models, which is shown in Figure \ref{ref:vis_texture}, while the other half are using random textures(paste random real images on the 3D models without any UV mapping). However, we do not use real textures in SMAL synthetic data generation. We have the following two reasons: (1) SMAL models do not have textures and generating high quality textures for SMAL models with fitting algorithms like SMALR \cite{Zuffi:CVPR:2018} requires a set of real images from comprehensive viewpoints which are hard to obtain in our case. (2) Table \ref{ref:tab_exp_texture} shows training on synthetic tiger with random textures only has significant better performance. This implies the "real" textures provided with CAD models are not realistic enough and still have large domain gap with real tiger's and tiger-like animals' textures although they are visually more similar to real than random textures. For amazing performance achieved by random textures, we think one possible main reason is the domain randomization give the model better generalization ability to other domains.  Another possible reason is that semantic parts of animals often have similar appearance which lessen the importance of realistic texture for part segmentation.

\section{Visualization for Tail Ambiguity}
\begin{figure*}[h]
\centering
\includegraphics[width=1\textwidth]{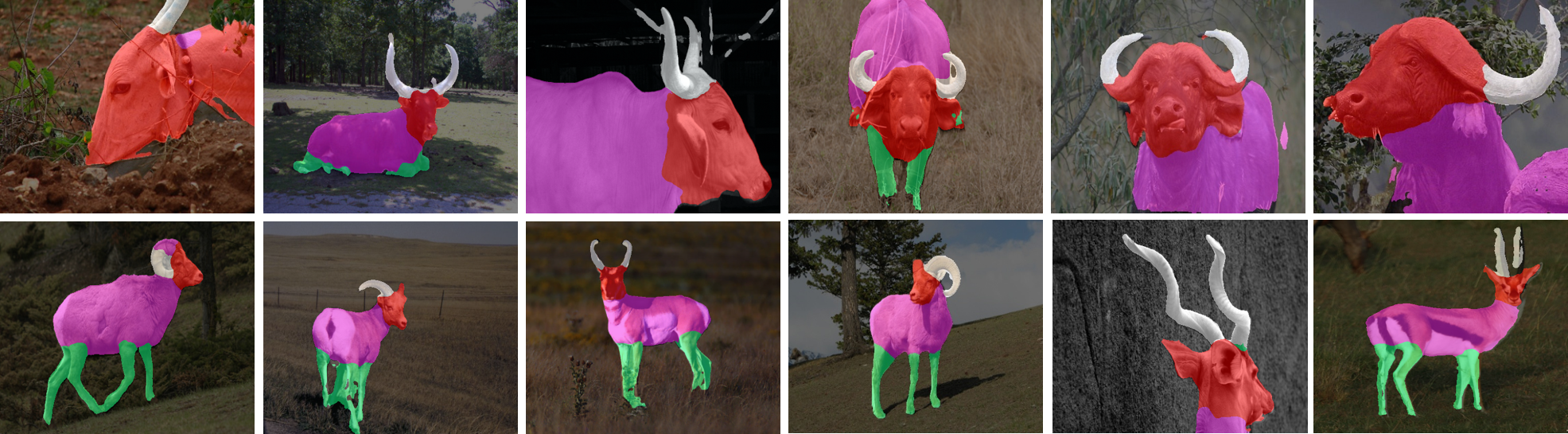}
\vspace{-4mm}
\caption{\textbf{Failure Examples in Classification of Tails and Horns(belonging to Head Class).}}
\label{ref:vis_tail_horn_ambiguity}
\end{figure*}

The tail ambiguity refers to the difficulty of distinguishing between tails and horns in our model (shown in Fig \ref{ref:vis_tail_horn_ambiguity}). In our case, horns should belong to animal head class while our models often predict them as tail. This tail ambiguity problem makes our Class-Balanced Pseudo-Label Re-Weighting (CB) approach fail to be applied for tail class since we can not be confident enough in these tail predictions. As our synthetic data does not contain animals with horns, it also indicates that the part segmentation model currently lacks the capability to handle unseen parts.

\section{Visualization of SMAL Fitting Results and Synthetic Data}
We have used more extreme viewpoints and larger range of camera distances than previous work \cite{mu2020learning,liu2022learning} when rendering the synthetic data. Furthermore, we use transformations (translation and rotation) to produce more position variations to reduce the domain gap between real and synthetic. The visualization is shown below in Fig \ref{ref:vis_SMAL}.\\

\noindent \textbf{Acknowledgement}
The authors gratefully acknowledge supports from ONR N00014-21-1-2690.

\begin{table}[]
\centering
\begin{minipage}[]{\linewidth}
\renewcommand\arraystretch{1.1}
\resizebox{\linewidth}{!}{
\begin{tabular}{c|cc|ccccc|c}
\multirow{2}{*}{method} & \multicolumn{2}{c|}{texture} & \multirow{2}{*}{head} & \multirow{2}{*}{torso} & \multirow{2}{*}{leg} & \multirow{2}{*}{tail} & \multirow{2}{*}{bg} & \multirow{2}{*}{mIoU} \\ \cline{2-3}
 & real & random &  &  &  &  &  &  \\ \hline
\multirow{2}{*}{SegFormer} & \xmark & \cmark & 71.44 & 46.94 & 35.01 & 22.29 & 83.86 & 51.91 \\
 & \cmark & \xmark & 57.69 & 26.65 & 28.42 & 24.36 & 78.49 & 43.12
\end{tabular}}
\caption{\textbf{Ablation of Animal Texture.} The real textures refer to training on 5000 CAD synthetic tiger images with the real textures provided with CAD models. The random textures refer to training on 5000 CAD synthetic tiger images with random textures from real images. The test set are tiger-like set. Numbers are averaged over 3 random seeds.}
\vspace{-4mm}
\label{ref:tab_exp_texture}
\end{minipage}
\end{table}

\begin{figure*}[h]
\vspace{-4mm}
\centering
\includegraphics[width=1\textwidth]{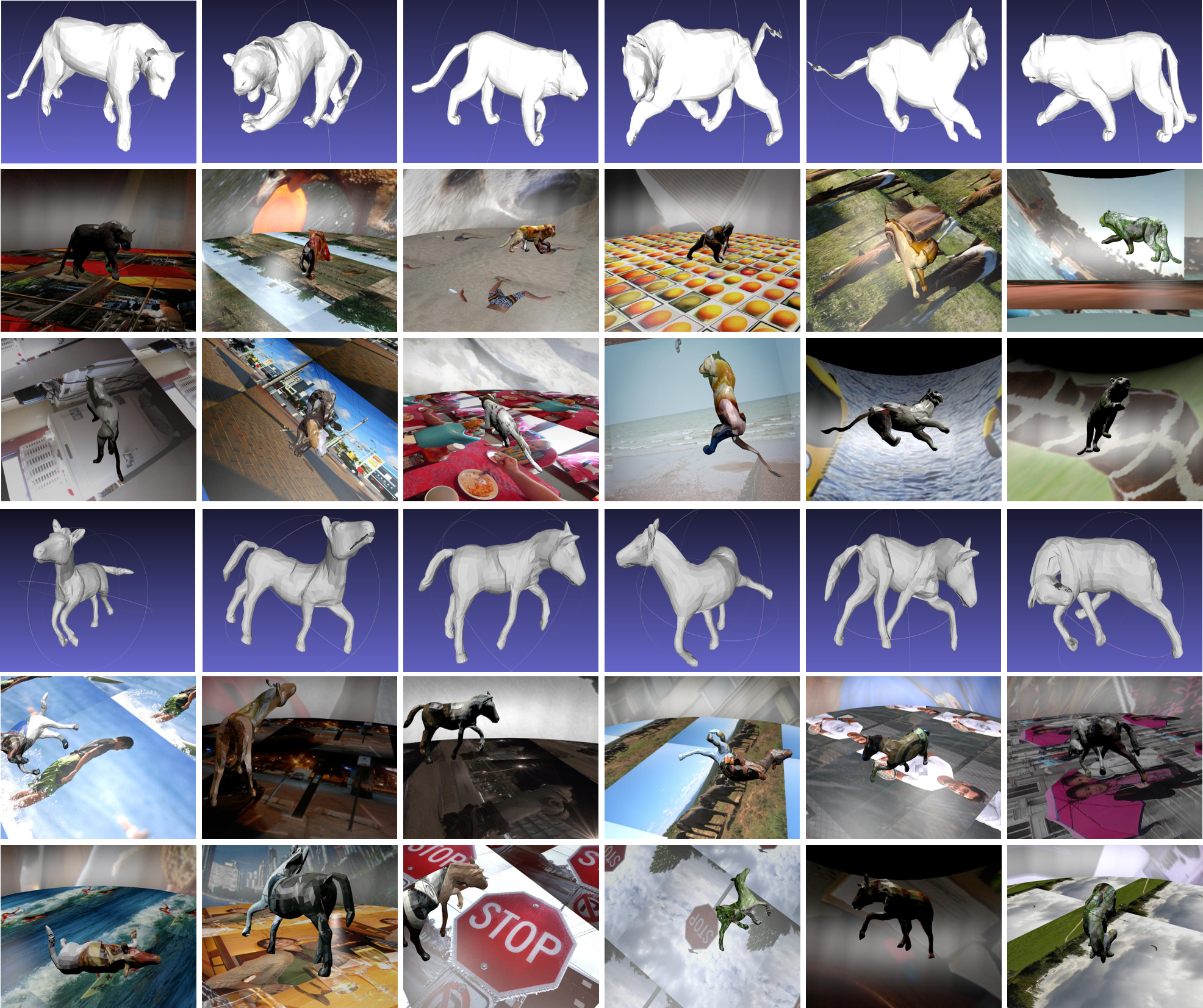}
\vspace{-4mm}
\caption{ \textbf{More SMAL fitting results and their corresponding SMAL synthetic Data.}}
\label{ref:vis_SMAL}
\end{figure*}

\end{document}